\documentclass[preprint,12pt]{elsarticle}

\usepackage{amssymb}
\usepackage{float}
\usepackage{hyperref}
\journal{Manufacturing Letters}

\begin{document}

\begin{frontmatter}

%% Title, authors and addresses

%% use the tnoteref command within \title for footnotes;
%% use the tnotetext command for theassociated footnote;
%% use the fnref command within \author or \address for footnotes;
%% use the fntext command for theassociated footnote;
%% use the corref command within \author for corresponding author footnotes;
%% use the cortext command for theassociated footnote;
%% use the ead command for the email address,
%% and the form \ead[url] for the home page:
%% \title{Title\tnoteref{label1}}
%% \tnotetext[label1]{}
%% \author{Name\corref{cor1}\fnref{label2}}
%% \ead{email address}
%% \ead[url]{home page}
%% \fntext[label2]{}
%% \cortext[cor1]{}
%% \affiliation{organization={},
%%             addressline={},
%%             city={},
%%             postcode={},
%%             state={},
%%             country={}}
%% \fntext[label3]{}

\title{Designing an LLM-Based Copilot for Manufacturing Equipment Selection}

%% use optional labels to link authors explicitly to addresses:
%% \author[label1,label2]{}
%% \affiliation[label1]{organization={},
%%             addressline={},
%%             city={},
%%             postcode={},
%%             state={},
%%             country={}}
%%
%% \affiliation[label2]{organization={},
%%             addressline={},
%%             city={},
%%             postcode={},
%%             state={},
%%             country={}}

\author[inst1]{Jonas Werheid}
\author[inst1]{Oleksandr Melnychuk}
\author[inst1]{Hans Zhou}
\author[inst1,inst3]{Meike Huber}
\author[inst2]{Christoph Rippe}
\author[inst2]{Dominik Joosten}
\author[inst2]{Zozan Keskin}
\author[inst3,inst5]{Max Wittstamm}
\author[inst3]{Sathya Subramani}
\author[inst3]{Benny Drescher}
\author[inst1]{Amon G\"oppert}
\author[inst1]{Anas Abdelrazeq}
\author[inst1,inst4]{Robert H. Schmitt}

% Affiliations
\affiliation[inst1]{organization={Laboratory for Machine Tools and Production Engineering, WZL, RWTH Aachen University},
            addressline={Campus-Boulevard 30}, 
            city={Aachen},
            postcode={52074}, 
            country={Germany}}

\affiliation[inst2]{organization={INC Innovation Center GmbH}, 
            addressline={Frankenthaler Str. 20}, 
            city={Munich}, 
            country={Germany}}

\affiliation[inst3]{organization={Hong Kong Industrial Artificial Intelligence \& Robotics Centre (FLAIR)},
            addressline={78 Tat Chee Avenue}, 
            city={Hong Kong}, 
            country={China}}

\affiliation[inst4]{organization={Fraunhofer IPT}, 
            addressline={Steinbachstr. 17}, 
            city={Aachen}, 
            postcode={52074}, 
            country={Germany}}

\affiliation[inst5]{organization={Corresponding Author},
            addressline={max.wittstamm@hkflair.org}}

%\corref{Corresponding Author: \texttt{max.wittstamm@hkflair.org}}}

%Hans feedback: Simpliuität, eine Figure, Weniger Use-case description, Wir wenden schnell/easy etwas an 
\begin{abstract}
%context and relevance
Effective decision-making in automation equipment selection is critical for reducing ramp-up time and maintaining production quality, especially in the face of increasing product variation and market demands.
%problem 
However, limited expertise and resource constraints often result in inefficiencies during the ramp-up phase when new products are integrated into production lines.
%gap
Existing methods often lack structured and tailored solutions to support automation engineers in reducing ramp-up time, leading to compromises in quality.
%question
This research investigates whether large-language models (LLMs), combined with Retrieval-Augmented Generation (RAG), can assist in streamlining equipment selection in ramp-up planning.
%approach
We propose a factual-driven copilot integrating LLMs with structured and semi-structured knowledge retrieval for three component types (robots, feeders and vision systems), providing a guided and traceable state-machine process  for decision-making in automation equipment selection.
%results
The system was demonstrated to an industrial partner, who tested it on three internal use-cases. Their feedback affirmed its capability to provide logical and actionable recommendations for automation equipment. 
%more specific
More specifically, among 22 equipment prompts analyzed, 19 involved selecting the correct equipment while considering most requirements, and in 6 cases, all requirements were fully met.

\end{abstract}

%%Graphical abstract
%\begin{graphicalabstract}
%\includegraphics{grabs}
%\end{graphicalabstract}

%%Research highlights
\begin{highlights}
\item Novel approach to reduce ramp-up time in manufacturing
\item Promising industrial feedback
\end{highlights}

\begin{keyword}
%% keywords here, in the form: keyword \sep keyword
 Manufacturing \sep Equipment Selection \sep  Generative AI \sep LLM \sep  Copilot
%% PACS codes here, in the form: \PACS code \sep code
%\PACS 0000 \sep 1111
%% MSC codes here, in the form: \MSC code \sep code
%% or \MSC[2008] code \sep code (2000 is the default)
%\MSC 0000 \sep 1111
\end{keyword}

\end{frontmatter}

\section{Introduction} 
%Context 
Global manufacturing demands agility, driven by increasing product complexity and rapidly changing market conditions, making efficient production ramp-up essential \cite{Magnanini2024}. Automation design, especially equipment selection, is critical in reducing ramp-up time and addressing challenges resulting from global trends like digitalization \cite{MuraPetruOvidiu2023DaEG}, globalization, and sustainability \cite{2668daa85fb440ddab4bd9029b5287bf}. 
%Relevance
The ramp-up phase relies on an effective design stage, where equipment choices must be made quickly to align with market demands and stay competitive. Key challenges mirror those in manufacturing, including complexity management \cite{faucris.241516043}-\cite{e26090747}, skills shortages \cite{MAGNANINI2021152}, supply chain reliability \cite{MELNYCHUK2022966}, and quality control \cite{MAGNANINI2021140}. 
While methods for equipment selection support have been introduced, they often rely on static rules and predefined knowledge \cite{park00}, which lack generalization and require users to adhere to specific syntax to align with the workflow \cite{bikas16}. 
%Gap
While Large Language Models (LLM) show potential to generate customized and dynamic interactions in other domains \cite{ji2023genreclargelanguagemodel}, they lack transparency \cite{Liao2024AI} and resilience \cite{10.1007/978-3-031-74186-9_17}-\cite{Hacker2023SustainableAR} and often suggest generic options \cite{Sarker_2024}.

For equipment selection, however, tailored \cite{9266587}, transparent \cite{Hollanek2020}, as well as dynamic solutions are required to support automation design and enable efficient ramp-up processes \cite{LEBERRUYER2023103877}-\cite{COLLEDANI2018197}. Retrieval-augmented generation (RAG) techniques have demonstrated their effectiveness in incorporating tailored information, reducing hallucinations, and improving response quality, especially within specialized domains. \cite{wang-etal-2024-searching}.
%Question: 
This article explores whether these recent advancements in factual-driven RAG-LLMs, combined with a state-machine architecture, can fulfill the requirements of the equipment selection process and guide automation equipment design effectively.

%Basic Literature / SOTA: 
The literature's existing approaches were investigated using the Scopus and Web of Science search platforms, which retrieved 198 and 85 articles, based on the selected search string provided in the Appendix (Section \ref{appendix:search_string}). Among the retrieved literature, Meyer et al. \cite{10710806} highlight the potential of generative AI for creating assembly instructions, demonstrating how LLMs can document processes to guide new workers on manufacturing lines. Similarly, Lim et al. \cite{10711843} investigate the use of LLMs for human-robot communication, where the LLM serves as an interface to the operational capabilities of robotic systems in manufacturing. Also, fine-tuned LLMs have been proposed as question-and-answer systems in product management and production line operations \cite{10704843}, with some utilizing RAG for enhanced information retrieval \cite{10649905}. These models have also been applied to extract information from technical manuals and other documents \cite{10706469}. Furthermore, LLMs have been explored for generating theoretical process selections, such as flowcharts \cite{10673464}.

However, no studies have been identified that propose LLM-driven methods or tools specifically designed to support automation engineers in equipment selection during the ramp-up phase.
%Method/Approach:
To address this gap, we propose a factual-driven copilot based on RAG-LLMs designed to support the selection of automation equipment in manufacturing.

%Structure:
First, the copilot design is introduced in Section \ref{copilot_design:copilot_design}. Next, Section \ref{case_study:case_study} presents feedback from an industrial partner. Lastly, limitations and potential directions for future research are derived in Section \ref{conclusion:conclusion}. 

\section{Copilot Design} \label{copilot_design:copilot_design}%INC / WZL 

\begin{figure}[H]
    \centering
    \includegraphics[width=1\linewidth]{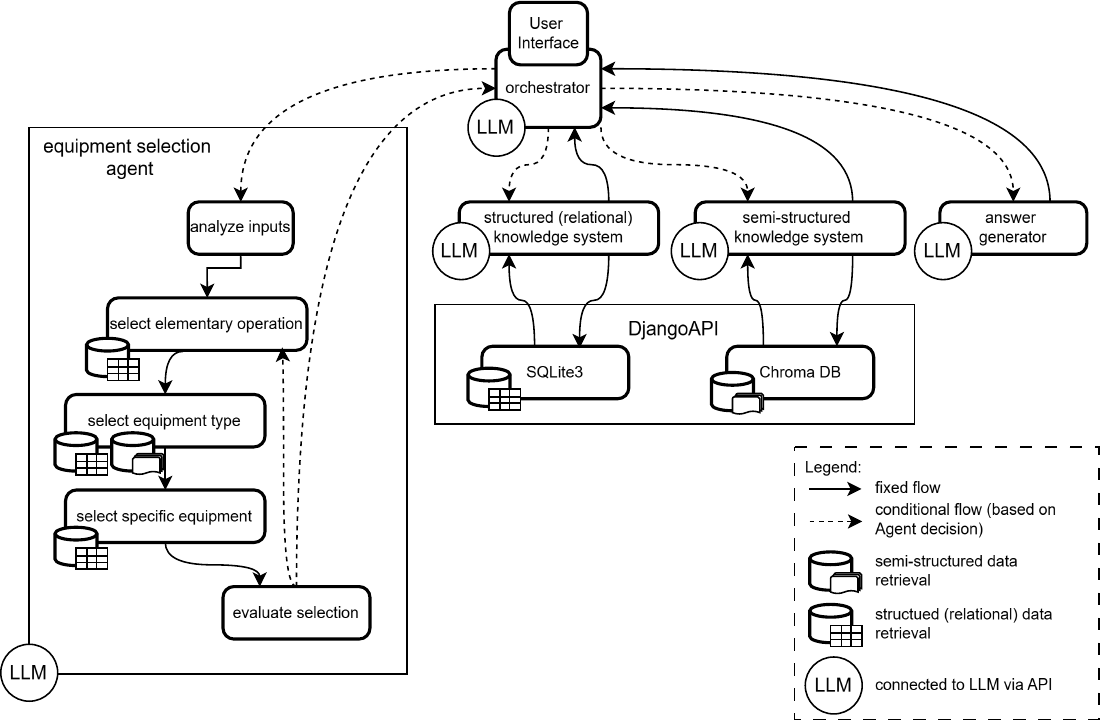}
    \caption{State machine diagram of the copilot}
    \label{fig:graph} 
    %Wording: equipment selection procedure, relational knowledge system, semi-structured knowledge system, answer generator procedure
\end{figure}

The LLM-based copilot consists of multiple components. At its core is the primary agent, which orchestrates subcomponents based on the multi-agent design pattern proposed by Wu et al. \cite{wu2024autogen}. These include the equipment selection procedure, the relational knowledge system, the semi-structured knowledge system, and the answer generator. All are either based on or connected to an LLM via an Application Programming Interface (API), with the knowledge behind the retrieval systems derived from scientific literature and lecture materials. Figure \ref{fig:graph} represents these components as a state machine, illustrating the copilot's interconnections.

The knowledge underpinning these systems is derived from a combination of scientific literature, lecture materials, and domain-specific datasets. 

Users can either ask general questions, which are answered by the primary agents LLM (orchestrator) with support from the relational and semi-structured knowledge systems, or initiate the equipment selection procedure, which also utilizes both systems in a multistep procedure. Firstly, the process of equipment selection is outlined, following a description of the semi-structured and structured knowledge systems.

The equipment selection process is triggered when the user specifies requirements for an assembly process. The LLM interprets the term specific based on a predefined prompt (pre-prompt) but can also be reinforced through a provided template. The workflow \ref{fig:graph} is initiated by analyzing the provided requirements. The requirements are structured and grouped into general assembly components, such as robots, and feeding systems based on a pre-prompt that describes them. Using this information, the system determines the appropriate elementary operation e.g. for robotic handling based on the structured knowledge. Following the determined elementary operation, a more specific type of equipment, such as a Cartesian Robot, is selected based on the structured and semi-structured knowledge. Therefore, the selection process relies on information from both relational and semi-structured knowledge.
This leads to specific equipment within the chosen elementary operation
category. Once a specific type of equipment is selected, the workflow progresses to the evaluation selection step, where the LLM reflects the suitability of the decision to alleviate Hallucination \cite{ji-etal-2023-towards}. In case of an unsuitable selection, additional information about the decision is requested from the user to enhance a further iteration of the equipment selection process based on the reflection, restarting the selection of the elementary operation. 

Alternatively, when tasked with answering general queries, the agent accesses the relational- and semi-structured knowledge systems to directly process the user request to provide a dedicated expert answer. The semi-structured knowledge system is a RAG-based component that processes text chunks containing semi-structured information on our partner's most relevant automation equipment types: Robotic systems, feeder technologies, and machine vision systems. This information is sourced from lecture slides and dissertations provided by the WZL institute and formatted in Markdown. The text chunks are embedded and indexed in RAG, with a chunk size of 750 and an overlap of 150. Overlapping chunks improve retrieval accuracy by ensuring context-aware responses \cite{church24}. The system retrieves the three most matching chunks, ordered by their relevance score. It is calculated as the dot product indicating the similarity between the embedding of the request and chunk \cite{Svec22}.

The structured knowledge system maps specific equipment such as feeders, robots, and vision systems to their corresponding attributes in a relational format. It is linked to product- and equipment catalogs by the selected supplier of the respective domains. The mapping is done based on scientific books \cite{boothroyd2005assembly}. 
Once the relevant information is retrieved, the system passes it to the answer generator, which synthesizes a response with a defined pre-prompt. 

The copilot is implemented in LangGraph \cite{LangGraph2024} using a state machine model. OpenAI's GPT-4o model \cite{OpenAI2024} is employed as the LLM model and integrated with LangGraph via its API. Django REST Framework provides the relational and vectorized database API \cite{django_rest}. Chroma \cite{ChromaDB} is used as a vector database, providing chunk embeddings generated with the open-source model all-MiniLM-L6-v2 by Hugging Face \cite{allMiniLM-L6-v2}. SQLite3 is selected to provide the structured knowledge. The user interaction is facilitated through Gradio, which visualizes the chat and the states \cite{Gradio2019}. The system is hosted directly on Microsoft Azure, enabling the industrial partner to access it for the case study \cite{MicrosoftAzure}.
\section{Industrial Case-study} \label{case_study:case_study}

The copilot, accessible as a web application, was presented to our industrial partner, an international manufacturer of plastic products with about 850 employees, operating six factories across three continents. Two automation engineers tested the system on three ramp-up planning use-cases: two for products in Europe and one in China. More specifically, the system was used to derive automation components for robots, feeders, and vision applications within the design process. Operational requirements were provided as context, and the components suggested by the copilot were compared against the internal selection. Table \ref{table:components_suggestions} provides examples of each component type, including the input prompt und equipment suggestions. However, not all prompts can be shown because they involve confidential company information.

\renewcommand{\arraystretch}{1.4}
\begin{table}[H]
\centering
\caption{Input prompts and components suggestions by partner and copilot for different operations}
\label{table:components_suggestions}
\resizebox{\textwidth}{!}{%
\begin{tabular}{p{0.12\textwidth}|p{0.10\textwidth}|p{0.28\textwidth}|p{0.20\textwidth}|p{0.28\textwidth}}
\textbf{Name} & \textbf{Type} & \textbf{Context (and prompt)} & \textbf{Component suggested by partner} & \textbf{Component suggested by copilot} \\
\hline
Pre-feeding of bushings & Feeder & The first equipment serves as an input buffer for 25 x 12 x 12 mm bushings weighing 20g, manually loaded by personnel. It acts as a 1-hour buffer, supplying at least 8 bushings every 40 seconds for subsequent processing. & SLL 800-1200 & Hudson Robotics PlateCrane EX, RNA SLK-05 \\
\hline
Movement to cleaning and inspection & Robot & Equipment is needed to work with a composite plastic and steel housing component, measuring 45 × 45 × 30 mm and weighing 60 g. It must provide a wide range of motion to perform complex tasks, including flexible movements and turning the housing to work on all its faces. Precision and repeatability are critical requirements. & Epson Articulated arm robots N2-A450 & Brand: RNA Type: Flexible Feeder Model: FlexCube 380, Brand: Epson Type: Articulated Arm Robot Model: C8-A1401 (C8XL), Brand: LMI Technologies Type: Laser Scanner Model: GOCATOR 2150-3R-REVD \\
\hline
Inspection & Vision & Equipment is needed for a visual inspection of the part afterward. The camera requires at least 5 Mega Pixels and two pictures need to be taken within a small timeframe. The field of inspection varies for each part. & Area-scan cameras with over 5MP/30FPS & OPTRONIS CP-CYCLONE-21-230-C \\
\end{tabular}%
}
\end{table}

Among the 22 equipment prompts analyzed, three involved selecting the correct equipment type, such as a feeder system, but not identifying the appropriate subtype, like a vibratory bowl feeder. In thirteen instances, a sufficient component was chosen by selecting the correct equipment subtype, such as a SCARA-Robot, and also considering most specific requirements, such as arm length. Notably, the selected component was fully suitable in six cases, successfully meeting all relevant requirements, reflecting a positive outcome. Therefore, the system delivered logical component suggestions, aligned with the partner's requirements. Nevertheless, it also had certain limitations. Firstly, it does not support layout design as an important aspect of the work of the automation engineer that influences the equipment selection itself. Additionally, it does not incorporate the ramp-up implementation phase. Nevertheless, it effectively supported the selection of appropriate automation equipment based on the provided context prompts.

\section{Conclusion} \label{conclusion:conclusion}  

%Restate the problem \\
The increasing complexity of manufacturing workflows, combined with a shortage of expertise, presents significant challenges in accelerating the integration of new products into production lines without compromising quality.
%Summary paper \\
This study introduced a factual-driven copilot utilizing RAG and tailored domain knowledge to assist engineers in streamlining ramp-up processes. The copilot demonstrated promising feedback in an industrial manufacturing use-case by providing structured and tailored features in traceable recommendations. These were derived from academic lectures, industrial literature, and a comprehensive supplier database, ensuring dependable local and international equipment suggestions. This underscores the potential of LLM-based systems to enhance the automation equipment selection process.
%Limitations, and Future research suggestions \\
However, limitations remain, as the copilot does not address the ramp-up implementation phase or layout design considerations. Future research should explore further integration, creating a comprehensive solution that spans design through to the ramp-up implementation phase.

\section{Usage Notes}
\label{usage}

The software will be made available on request to the authors Jonas Werheid (jonas.werheid@rwth-aachen.de) and Dominik Joosten (dominik.joosten@rwth-aachen.de). %It is licensed under GNU General Public License v3.0. 

%The dataset generated for this research is accessible on INSERT HERE via DOI \href{https://example.com}{(\url{https://example.com})}. The dataset is licensed under for example the Creative Commons Attribution 4.0 International License (CC BY 4.0). The developed software is partly available on INSERT HERE \href{https://example.com}{(\url{https://example.com})} and licensed under GNU Affero General Public License v3.0 for example.

\section{Declarations}
\subsection{Funding} 
Our research is funded by AIR@InnoHK research cluster of the Innovation and Technology Commission (ITC) of the HKSAR Government. The results presented in this paper within the scope of the research project '1-5 Ramp-Up Time Reduction by Learning from Past Process Cycles' has come to a successful completion because of the support from ITC. We would like to express our sincere gratitude to them.  
%We authors would like to thank the Hong Kong Industrial Artificial Intelligence \& Robotics Centre (FLAIR) for funding this research. 
\subsection{Competing interests}
The authors have no competing interests to declare that are relevant to the content of this article.
\subsection{Author Contributions}
\textbf{Conceptualization, Writing – original draft, Writing – review and editing:} Jonas Werheid, Oleksandr Melnychuk, Hans Zhou, Meike Huber, Christoph Rippe, Dominik Joosten, Zozan Keskin. \\ \textbf{Conceptualization, Writing – review and editing:} Max Wittstamm, Sathya Subramani, Benny Drescher, Amon G\"oppert, Anas Abdelrazeq, Robert H. Schmitt.

\appendix
\section{Search String}
\label{appendix:search_string}
\noindent\parbox{\linewidth}{
\texttt{
(("manufacturing" OR "production line*" OR "production planning" OR "production optimization" OR "production processes" OR "automation engineering" OR "fabrication" OR "assembly line*" OR "assembly automation" OR "industrial automation") AND\\
("ramp-up" OR "ramp up" OR "quality" OR "efficiency" OR "resources" OR "optimization" OR "quality assurance" OR "quality control" OR "labor") AND\\
("large-language models" OR "language models" OR "LLM*" OR "retrieval-augmented generation" OR "RAG" OR "GPT*" OR "copilot" OR "generative AI" OR "AI assistant" OR "intelligent assistant"))
}}

%% If you have bibdatabase file and want bibtex to generate the
%% bibitems, please use
%%
 \bibliographystyle{elsarticle-num} 
 \bibliography{cas-refs}

%% else use the following coding to input the bibitems directly in the
%% TeX file.

% \begin{thebibliography}{00}

% %% \bibitem{label}
% %% Text of bibliographic item

% \bibitem{}

% \end{thebibliography}
\end{document}